\documentclass[sigconf]{acmart}

\usepackage{graphicx}
\usepackage{epstopdf}
\usepackage{booktabs} 
 \usepackage{amsmath}

\usepackage{amsmath}
\usepackage{bm}

\usepackage{multirow}
\usepackage{indentfirst}
\usepackage{subfigure}
\usepackage{epsfig}
\usepackage{epstopdf}
\usepackage{xspace}
\usepackage{pdfpages}
\usepackage{makecell,rotating}

\usepackage{dsfont}

\usepackage{subeqnarray}

\usepackage{algorithm}
\usepackage{algorithmic}
\usepackage[toc,page]{appendix}
\usepackage{hyperref}

\newcommand{\ignore}[1]{}

\ignore{
\algnewcommand\algorithmicprocedure{\textbf{procedure}}
\algnewcommand\PROCEDURE{\item[\algorithmicprocedure]}%
\algnewcommand\algorithmicendprocedure{\textbf{end procedure}}
\algnewcommand\ENDPROCEDURE{\item[\algorithmicendprocedure]}%
\algnewcommand{\algvar}[1]{{\text{\ttfamily\detokenize{#1}}}}
\algnewcommand{\algarg}[1]{{\text{\ttfamily\itshape\detokenize{#1}}}}
\algnewcommand{\algproc}[1]{{\text{\ttfamily\detokenize{#1}}}}
\algnewcommand{\algassign}{\leftarrow}
}

\setcopyright{rightsretained}

\def\BibTeX{{\rm B\kern-.05em{\sc i\kern-.025em b}\kern-.08emT\kern-.1667em\lower.7ex\hbox{E}\kern-.125emX}}

\copyrightyear{2018}
\acmYear{2018}
\setcopyright{acmlicensed}
\acmConference[KDD '19]{KDD '19: ACM SIGKDD Conference on Knowledge Discovery and Data Mining}{August 04--08, 2019}{}
\acmBooktitle{KDD '19: ACM SIGKDD Conference on Knowledge Discovery and Data Mining, August 04 -- 8, 2019, Anchorage, Alaska USA}
\acmPrice{15.00}
\acmDOI{10.1145/1122445.1122456}
\acmISBN{978-1-4503-9999-9/18/06}

\renewcommand\footnotetextcopyrightpermission[1]{} 
\begin{document}

%
\title{Interpreting Deep Learning Model Using Rule-based Method}


\author{Xiaojian Wang}
\affiliation{%
  \institution{Purdue University}
  }
\email{wangxjbuaa@foxmail.com}

\author{Jingyuan Wang}
\affiliation{%
  \institution{Beihang University}
  }
\email{jywang@buaa.edu.cn}

\author{Ke Tang}
\affiliation{%
  \institution{Tsinghua University}
  }
\email{ketang@tsinghua.edu.cn}
%

%
%
\begin{abstract}
Deep learning models are favored in many research and industry areas and have reached the accuracy of approximating or even surpassing human level. However they've long been considered by researchers as black-box models for their complicated nonlinear property. In this paper, we propose a multi-level decision framework to provide comprehensive interpretation for the deep neural network model. 

In this multi-level decision framework, by fitting decision trees for each neuron and aggregate them together, a multi-level decision structure (MLD) is constructed at first, which can approximate the performance of the target neural network model with high efficiency and high fidelity. In terms of local explanation for sample, two algorithms are proposed based on MLD structure: forward decision generation algorithm for providing sample decisions, and backward rule induction algorithm for extracting sample rule-mapping recursively. For global explanation, frequency-based and out-of-bag based methods are proposed to extract important features in the neural network decision. Furthermore, experiments on the MNIST and National Free Pre-Pregnancy Check-up (NFPC) dataset are carried out to demonstrate the effectiveness and interpretability of MLD framework. In the evaluation process, both functionally-grounded and human-grounded methods are used to ensure credibility.

\end{abstract}

%
%
%
%
%


%

%
\maketitle

\section{Introduction}

Deep learning models play an increasingly significant role in real world applications, such as image and speech recognition, medical diagnostics and intelligent recommendation.  In spite of the incredible performance, there's one critical issue with deep learning models that we cannot ignore: they are generally considered by researchers as black-box models which limits their application in the areas such as causal inference, trust (i.e., in healthcare applications, a trusting model which can be understood, validated and edited is needed for the experts \cite{caruana2015intelligible}), fairness (i.e. internal prejudice of the training data might be amplified during the model training \cite{dwork2012fairness}) and some security-focused scenarios (i.e., it's hard to detect the adversarial attacks in some image recognition tasks\cite{Akhtar2018threat}). 

It's worth noticing that more and more efforts have been devoted to make the deep learning models interpretable. One of the ideas is transforming the neural network decision into a set of rule mappings. The rule extraction techniques can be divided into two categories: decompositional approaches and pedagogical approaches \cite{andrews1995survey}. Some of the earliest work adopted the decompositional approach to extract the boolean rules at the individual level. The basic idea of the decompositional approach is to treat the rule extraction as a searching process. The main defect of the decompositional methods is the complexity of the search process is exponential with the number of neurons and layers. 

Compared to the decompositional approaches which search and analyze the structure of neural networks, the pedagogical approaches treat the neural network as a whole and  aim to extract the rules that replace the original network as accurately as possible. But the applicability of the pedagogical methods is also limited, because in the context of deep networks, we need more complex rule sets to get the performance approximate to neural network and that will cause the structures generated by pedagogical methods to be extremely complicated.

To address the limitations of existing rule-based methods, we propose an interpretable framework for the multi-layer neural network based on a multi-level decision structure. At first, a multi-level decision structure (MLD) is derived by transforming the discretized activation function of each neuron into a decision tree and linking the generated trees layer by layer. Different from the other pedagogical rule-based methods, MLD preserves the original structure of the neural network. 

Then, based on the MLD structure we propose forward decision generation algorithm and reverse rule induction algorithm to produce local explanation for each prediction. Applying the forward decision generation algorithm, we can generate the decision of the MLD layer-by-layer for any given input sample, due to the high fidelity of MLD, it can approximate the performance of the original neural network in high accuracy. Applying the reverse rule induction algorithm which merges rules retrospectively, we can finally get the rule set which is corresponding to the input space for any given sample.  

In order to extract global explanations which can further help researchers interpret the overall reasoning of the model, we design frequency and out-of-bag methods for evaluating the importance of input features. Furthermore, We evaluate the interpretability of MLD on MNIST and NFPC dataset, predictivity and fidelity are compared with a few baselines, visual and global explanations are given and validated as well.

To summarize, the major contributions of this paper are the following:
\begin{itemize}
\item{We propose multi-level decision (MLD) framework. To the best of our knowledge, compare to other rule-based approaches, MLD combines the advantages of both decompositional and pedagogical methods, achieves higher fidelity and meanwhile preserves the original network structure.}

\item{MLD framework can not only provide local explanations for a given sample by generating rule mapping, but also provide global explanations in the sense of feature importance.}

\item{Extensive experiments on two different tasks with both public and large-scale real-world dataset show the effectiveness of the proposed MLD approach. Interpretability is evaluated by both functionally-grounded and human-grounded methods.}
\end{itemize}

The rest of the paper is organized as follows. In Section 2, we survey the works related to interpretable deep learning. Section 3 introduces the preliminaries, problem definition and technical details of our proposed method. In Section 4, we empirically evaluate the performance of our model on various data. Finally, Section 5 is the conclusion.

\section{Related Works}

Interpretability methods for deep learning aim at providing understandable explanation towards the black-box nature of deep learning models. With reference to the research by Benn \textit{et al.} \cite{kim2017interpretable} and a slight adjustment, We further divide the interpretability methods of deep learning models into four 
categories by : 1) hidden-layer investigation methods, 2) sensitivity and gradient-based methods, 3) mimic and surrogate model methods, 4) interpretable deep learning methods.

The hidden-layer investigation methods implement the interpretation of deep learning models by analyzing or visualizing the behavior of hidden layers. In the research of Matthew D. \textit{et al.} \cite{zeiler2014visualizing}, a deconvolutional network is used to perform the mapping from the feature activities to the input pixel space.  Karpathy \textit{et al.}  \cite{karpathy2015visualizing} extended this idea to the recurrent neural networks. Alternatively, Bau \textit{et al.}  \cite{bau2017network} proposes a network dissection framework for quantifying the semantic representation of CNN by aligning the hidden units to a sets of concepts. Besides discovering the knowledge learned by the network, hidden layer methods are also implemented to study the training process of the network \cite{yosinski2014transferable,alain2016understanding}.

Sensitivity and gradient-based methods provide explanation by measuring how the change of variables or weights affects the output. Garson \textit{et al.} \cite{Garson1991} proposed an algorithm by calculating variable contribution based on the absolute values of connection weights. Based on the work of Garson, Olden \textit{et al.} \cite{olden2002illuminating} studied the significance of variables and weights using a sampling method. Some of other sensitivity methods investigate the neural network by perturbing the test point or by fitting a simpler model locally \cite{simonyan2013deep,li2016understanding,koh2017understanding} .

More closely related to our work, mimic and surrogate model methods use some more interpretable models to simulate the structure or behavior of a neural network to provide explanations. There're two categories of these methods: 1) \textit{Linearization methods}, which aim to establish a linear mapping between the input and output space. Such as relevance propagation \cite{bach2015pixel} and deep taylor decomposition \cite{montavon2017explaining}. 2) \textit{Rule-based methods}, which demonstrate their interpretability by extracting symbol rules from neural networks. Rule-based methods can be further divided into decompositional approaches and pedagogical approaches \cite{andrews1995survey}. Decompositional methods treat the problem of extracting rules from neural networks as a search process. e.g. extracting M-of-N rules from neural network based on clustering methods \cite{towell1992interpretation}, extracting rules from pruned feedforward neural network \cite{setiono1995understanding, setiono1997neurolinear}, and searching rules by finding the combinations \cite{krishnan1999search}. The biggest challenge for decomposition methods is as the number of neurons increases, these methods will face the explosion of rule combinations. Therefore, the decompositional methods usually reduce the computational complexity by some pruning or aggregation methods. In contrast to the decompositional methods, pedagogical methods treat the neural network as a unified whole, and aim to generate rules which can better approximate the performance of original network. e.g. RF and RN method \cite{saito2002extracting}, TREPAN \cite{craven1996extracting,trepan2}, tree construction methods based on genetic algorithm \cite{krishnan1999search} and sampling \cite{bastani2017interpreting}.

There are also some studies dedicated to building interpretable deep learning methods. e.g. Capsule \cite{sabour2017dynamic} method can produce output vectors whose length represent the probability that the entity exists and the orientation for the instantiation parameters. Based on the mechanism of attention, visual attention \cite{allport1989visual,xu2015show} can be used to visualize the focus of the the network. 

\section{Methodology}

\subsection{Preliminaries}

This section introduces the notations that will be used throughout the paper.

The interpretability method we proposed focuses on multi-layer perceptron model (MLP). Considering a trained MLP with $L$ layers for a multiclass classification problem, Let $\mathcal{X}\in {\rm I\!R}^d$ denote the input space and $\mathcal{Y}=[K]$ denote the output space. Suppose there're $m^{(l)}$ nodes in $lth$ layer of the MLP, $\mathbf{w}^{(l)}$ be the trained weight between $lth$ and $(l+1)th$ layer and $a$ be the activation function, For the training set $\mathcal{D}=\{\mathbf{x}^{(1)}_n,\mathbf{y_n}\}^N_{n=1}$ of size N, where $\mathbf{x}^{(1)}_n = (x^{(1)}_{n,1},...,x^{(1)}_{n,d})\in \mathcal{X}$ is feature vectors of length d and $y_n \in \mathcal{Y}$. Let $X^{(l)}=(\mathbf{x}_1^{(l)},\mathbf{x}_2^{(l)},..,\mathbf{x}_N^{(l)})$ be the output of the $lth$ layer of the MLP, then each of the $\mathbf{x}^{(l)}_i$ should be a vector with length $m^{(l)}$. Generally we have:
\begin{equation}
    x^{(l)}_{i,k} = a(\mathbf{w}^{(l-1)}_k\mathbf{x}^{(l-1)}_i), \forall 1\le i\le N, 1\le k\le m^{(l)}, 2\le l\le L
\end{equation}

Note that, for convenience we only consider the MLP models which target for classification problems and use ranged activation function such as tanh/sigmoid.

\subsection{Problem Definition}
For a general classification task which uses MLP in the modeling process, given a train dataset $\mathcal{D}=\{\mathbf{x}^{(1)}_n,\mathbf{y_n}\}^N_{n=1}$ we can then apply some gradient optimization methods to train a MLP. Besides analyzing the performance of the model prediction, we may also concern about how the model derive the prediction. Generally, the related interpretability problems can be divided into two aspects:

1) Local explanation: local explanation focuses on knowing the reasoning process for a certain decision. Inspired by some linearity methods and rule-based methods, in the context of local explanation, we aim to find a mapping function $\hat f(\mathbf{x})$ from the input space and the output space which is much easier to be understood than the mapping function $f(\mathbf{x})$ of the original MLP.
\begin{equation}
 \hat f(\mathbf{x}) \approx f(\mathbf{x})
\end{equation}

More specifically, in this paper we focus on the rule-based method in which $\hat f(\mathbf{x})$ can be represented by a set of rule mappings .

2) Global explanation: global explanation implies knowing what patterns are present in general. In this paper, our goal for the global explanation is to determine the key features which put most impact on the final decision.

\subsection{Framework for Interpretation Process Based on Multi-level Decision Structure(MLD)}

\begin{figure}[t]
  \centering
  \includegraphics[width=\columnwidth]{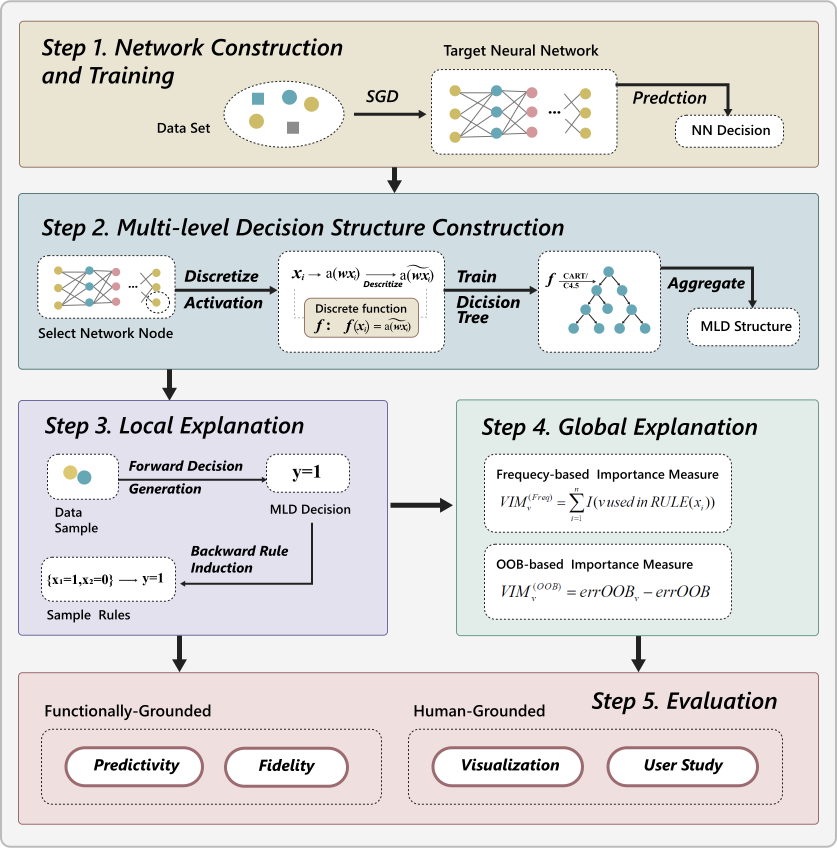}
  \caption{Framework for Interpretation Process Based on Multi-level Decision Structure(MLD)}
  \label{fig:framework}
\end{figure}

As shown in \textbf{Figure \ref{fig:framework}}, framework for interpretation process based on multi-level decision structure can be split into five steps:
\begin{enumerate}
    \item \textbf{Network construction and training}. First and foremost, for a classification task based on a given dataset, we can apply stochastic gradient descent (or other network training methods) to train a MLP. In the following steps, we will use this MLP as the target neural network for interpretaion.
    \item \textbf{Multi-level decision structure construction}.  In this step, we build the multi-level decision structure(MLD) by fitting the activation function of each node in the target network to a decision tree, then we aggregate them together.
    \item \textbf{Local explanation}. As a rule-based approach, MLD framework provides the local explanation for a given sample by transforming the sample decision into a rule mapping. Based on the MLD structure generated in \textbf{step (2)}, we proposed the forward decision generation algorithm to generate the sample decision and backward rule induction algorithm to derive the decision rules recursively. 
    \item \textbf{Global explanation}. In terms of global explanation, MLD framework aims at finding the important features in the neural network decision process. Based on the rule mapping derived in \textbf{step (3)}, frequency-based and out-of-bag importance measure methods are proposed.
    \item \textbf{Evaluation}. Two categories of evaluation methods for the interpretability are used in MLD framework: functional-grounded and human-grounded. In functional-grounded methods, predictivity and fidelity scores are measured according to the accuracy and F1 score of the MLD decisions regarding to the original data labels and neural network predicted labels. In human-grounded methods, we provide visual explanations and conduct user study as well.
    
\end{enumerate}

\subsection{Construction of Multi-level Decision Structure}
\begin{figure}[h]
  \centering
  \includegraphics[width=\columnwidth]{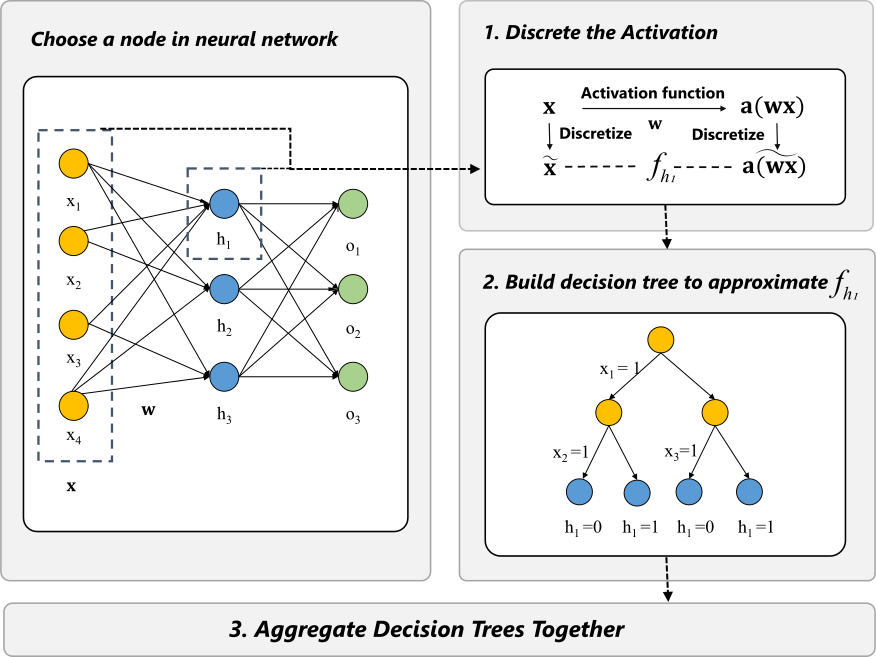}
  \caption{Steps for Construction of Multi-level Decision Structure (MLD)}
  \label{fig:discretize}
  \Description{}
\end{figure}

Multi-level decision structure here implies a hierarchical structure in which each layer contains several decision trees and the leaf nodes of these trees can be treated as the input feature of the next layer. As shown in \textbf{Figure \ref{fig:discretize}}, for a given data set $\mathcal{D} = \{\mathbf{x}_n^{(1)}, \mathbf{y}_n\}^N_{n=1}$ and the trained MLP, the generation process of MLD can be split into three steps.

1) Discretize the activation function for each neuron. For the $kth$ node in the $lth$ layer in a given MLP, we have:

\begin{equation}
    x^{(l+1)}_{i,k} = a(\mathbf{w}^{(l)}_k\mathbf{x}^{(l)}_i), \forall 1\le i\le N, l\le L-1
\end{equation}

In order to transform the activation function into a discrete function, we need to discrete both the input vectors and output vectors. Since the output value of activation functions like sigmoid and tanh are restricted in $\{0,1\}$. We here reduce them into boolean discrete function for simplicity. Note that the boolean mapping is not the only choice, other discretization method can also be applied as long as it is preferred.
 
\begin{equation}
    \tilde x^{(l)}_{i,k} = 
    \left\{
    \begin{array}{lr}
       1 & x^{(l)}_{i,k}>=0.5\\
       0 &   x^{(l)}_{i,k}<0.5
    \end{array}
   \right.
\end{equation}

Here we may encounter with situations where the input vectors $ \mathbf{\tilde x}^{(l)}_i,\mathbf{\tilde x}^{(l)}_j$ are the same but the output vactors $\tilde x^{(l+1)}_{i,k},\tilde x^{(l+1)}_{j,k}$ maybe different, which is not eligible for a function mapping. We further reduce the different output values into the mode of the output values which share the same input vectors in this situation.
\begin{equation}
    \tilde x^{(l+1)}_{i,k} = MODE\{\tilde x^{(l+1)}_{j,k} ~| ~ \forall ~ \mathbf{\tilde x}_j^{(l)} =  \mathbf{\tilde x}_i^{(l)}~ , j\in\{1,2,...,N\}\}
\end{equation}

Then for the $kth$ neuron in $(l+1)th$ layer, Based on the discretized input and output pairs $ \{\mathbf{\tilde {x}}^{(l)}_i, \tilde {x}^{(l+1)}_{i,k}\}_{i=1}^N$, a corresponding boolean function $f^{(l+1)}_k$ can be generated and represented as following:

\begin{equation}
    f^{(l+1)}_k(\mathbf{x}) = 
      \tilde x^{(l+1)}_{i,k},~~ if~ \mathbf{x}=\mathbf{\tilde x}^{(l)}_{i}
\end{equation}

2) For every given discretized function $f^{(l)}_k$, approximate it with a decision tree $T^{(l)}_k$ and optimize the error in the sample space. Here we can use any decision tree generation function to obtain the aproximation. CART algorithm, which uses gini as the criterion is prefered here.

3) The MLD structure can be obtained by treating the output of each decision tree as the input of the next layer. Here we denote the MLD as $\mathbf{T}$ which represents the collection of the decision trees $\mathbf{T}=\{T^{(l)}_k|\forall 1\le k \le m^{(l)}, 2\le l\le L\}$.

\subsection{Forward Decision Generation Algorithm}

Based on the MLD structure, a foward decision generation algorithm is proposed to derive the decision for a given sample. Suppose $T_k^{(l)}$ be the generated decision tree for $kth$ node in $lth$ layer, $T_k^{(l)}(\mathbf{x}^{(1)}_i)$ be the output value of the given original input $\mathbf{x}^{(1)}_i$. For the $L$ layers MLP, our goal here is generating $T^{(L)}_k(\mathbf{x}^{(1)}_i)$ for any $1\le k\le m^{(l)}$ and $2\le l\le L$. In particular, output for the decision trees in the last layer $\{T^{(L)}_k(\mathbf{x}^{(1)}_i)|1\le k\le m^{(L)}\}$ can be treated as the final output value for the MLD. 

Every $T^{(l)}_k(\mathbf{x}^{(1)}_i)$ can be generated layer by layer fowardly. For the corresponding decision tree of the $jth$ node in the second layer $T_j^{(2)}$, according to the numeric value of sample $\mathbf{x}^{(1)}_i$ in each dimension, $T_k^{(2)}(\mathbf{x}^{(1)}_i)$ can be generated by going through from the root of $T_k^{(2)}$ to its leaf node. Respectively, every $T_k^{(3)}(\mathbf{x}^{(1)}_i)$ can be derived by implementing  $(T_1^{(2)}(\mathbf{x}^{(1)}_i),T_2^{(2)}(\mathbf{x}^1_i)..,T_{m^2}^{(2)}(\mathbf{x}^{(1)}_i))$  into the tree structure of $T^{(3)}_k$. \textbf{Algorithm \ref{alg:fdg}} present the procedure for the forward decision generating process.

\begin{algorithm}[htb]
  \caption{Decision and Rule Generation for Single Decision Tree }
  \label{alg:singlerule}
  \begin{algorithmic}[1]
    \REQUIRE
     decision tree $T$; input features $\mathbf{x}$
    \ENSURE
     Rule set $R_\mathbf{x}$; decision result $T(\mathbf{x})$ 
    
    initialize $R_\mathbf{x}=\Phi$, $T(\mathbf{x}) =0$ 
    \WHILE{$T$ is not empty}
    \IF{$T$ has no branch}
        \STATE{$T(\mathbf{x})=T$}
    \ELSE
    \FORALL {branch $c$ in $T$ for decision variable $v$ and with critical value $c_v$  }
        \IF{c is discrete variable}
            \IF{$x_v == 0$}
                \STATE{$T =$  left chance node branching from $c$}
                \STATE{r = $(v,c_v,=)$}
            \ELSE
                \STATE{$T =$  right chance node branching from $c$}
                \STATE{r = $(v,c_v,\ne)$}
            \ENDIF
        \ELSE 
            \IF{ $x_{v} \le c_v$}
                \STATE{$T =$  left chance node branching from $c$}
                \STATE{r = $(v,c_v,\le)$}
            \ELSE
                \STATE{$T =$  right chance node branching from $c$}
                \STATE{r = $(v,c_v,>)$}
            \ENDIF        
        \ENDIF
        
        \STATE{$R_\mathbf{x}=R_\mathbf{x}\cup r$}
    \ENDFOR
    \ENDIF
    \ENDWHILE
  \end{algorithmic}
\end{algorithm}

\begin{algorithm}[htb]
  \caption{Forward Decision Generation}
  \label{alg:fdg}
  \begin{algorithmic}[1]
    \REQUIRE
     Input sample $\mathbf{x}^{(1)}_i$; MLD structure $\mathbf{T}$;
    \ENSURE
      Output of every decision tree for every given sample $\{T^{(l)}_k(\mathbf{x}^{(1)}_i)$ | $\forall1\le k\le m^{(l)},2\le l\le L\}$;
      
    \FOR{$l=1$ to $L$}
      \FOR{$k=1$ to $m^{(l)}$}
      \IF{$l==2$}
        \STATE{Input feature vector $(x^{(1)}_{i,1},x^{(1)}_{i,2},...,x^{(1)}_{i,d})$ and decision tree structure $T^{(2)}_k$, Apply \textbf{Algorithm \ref{alg:singlerule}} to compute $T^{(2)}_k(\mathbf{x}^{(1)}_i)$}
      \ELSE
      
        \STATE{ Input decision tree structure $T^{(l)}_k$ and feature vector
        
        $(T^{(l-1)}_1(\mathbf{x}^{(1)}_i),T^{(l-1)}_2(\mathbf{x}^{(1)}_i),...,T^{(l-1)}_{m^{(l)}}(\mathbf{x}^{(1)}_i))$,
        
        Apply \textbf{Algorithm \ref{alg:singlerule}} to compute $T^{(l)}_k(\mathbf{x}^{(1)}_i)$}
        \ENDIF
      \ENDFOR
    \ENDFOR
  \end{algorithmic}
\end{algorithm}

\begin{figure}[h]
  \centering
  \includegraphics[width=\columnwidth]{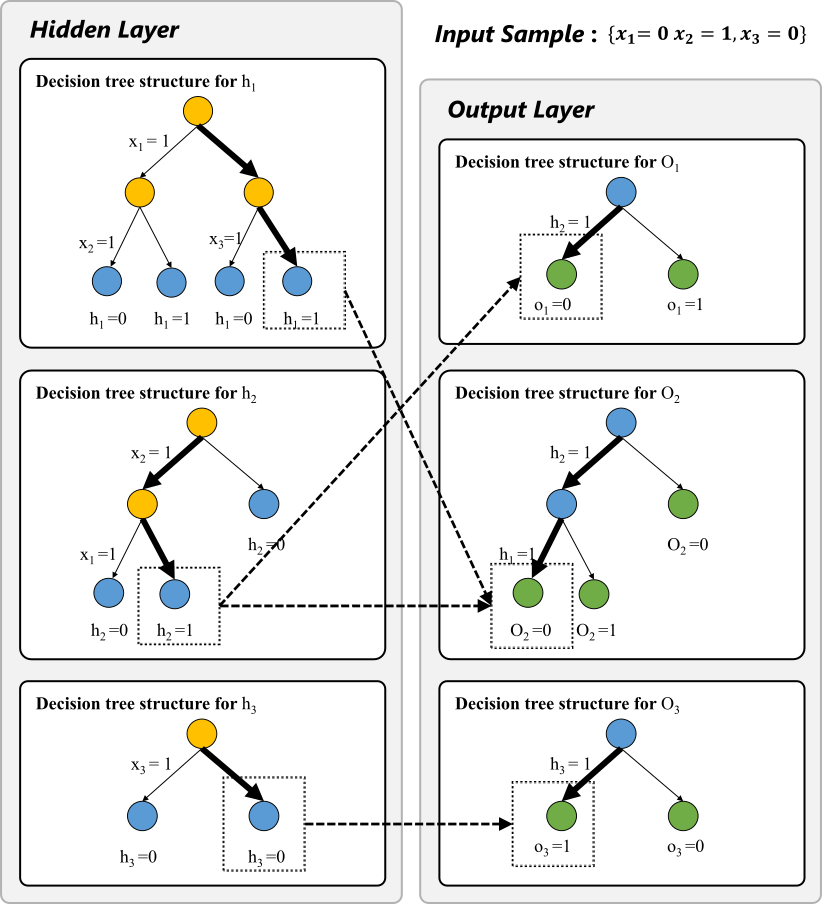}
  \caption{Demonstration for the Forward Decision Generation Process (Target MLP is the structure shown in Figure \ref{fig:discretize})}
  \label{fig:forward}
  \Description{Note that we still use the network structure in }
\end{figure}

\subsection{Backward Rule Induction Algorithm}
One objective of the proposed model is to generate the local explanation. More specifically, in the MLD scenario, we want to find a rule mapping  $R^{(L)}(\mathbf{x}^{(1)}_i)$ from input space to the output space for a given sample $\mathbf{x_i^{(1)}}$, For example: 
\begin{equation}
\begin{split}
    R_1^{(L)}(\mathbf{x^{(1)}_5}): (x^{(1)}_{5,1}=1)\land (x^{(1)}_{5,3}=1)\land (x^{(1)}_{5,5}=0)\\\implies (y_{5,1}=0)
\end{split}
\end{equation}

This rule mapping suggests for the $5th$ data sample, according to $x^{(1)}_{5,1}=1,x^{(1)}_{5,3}=1$ and $x^{(1)}_{5,5}=0$, we can conclude the final decision  $y_{5,1}=0$ , Through the rule induction process, such a mapping rule for a given sample can be derivated backwardly and  recursively. The main idea of the back rule induction algorithm is as follow:

According to the decision tree approximate method, we can get the MLD structure $\mathbf{T}=\{T^{(l)}_k|\forall 1\le k \le m^{(l)},2\le l\le L\}$ and the output decision of each decision tree $T^{(l)}_k(\mathbf{x}_i^{(1)})$ for any given input sample $\mathbf{x}^{(1)}_i$. Based on the decision making process of MLD, generation of decision for the $kth$ node in the $Lth$ layer $T^{(L)}_k(\mathbf{x}_i)$ can be reduced to a rule mapping:

\begin{equation}
\begin{split}
R^{(L)}_k(\mathbf{x}_i): (x^{(L-1)}_{i,v_1}=T^{(L-1)}_{v_1}(\mathbf{x}^{(1)}_i))\land(x^{(L-1)}_{i,v_2}=T^{(L-1)}_{v_2}(\mathbf{x}^{(1)}_i))\\
...\land(x^{(L-1)}_{i,v_m}=T^{(L-1)}_{v_m}(\mathbf{x}^{(1)}_i))  \implies (y_{i,k}=T^{(L)}_k(\mathbf{x}^{(1)}_i))
\end{split}
\end{equation}

Where $v_1,...v_m \in \{1,2,...,m^{(L-1)}\}$ stands for the index of the node in the $(L-1)th$ layer we use to produce the decision for $T^{(L)}_k(\mathbf{x}_i^{(1)})$ in the branching procedure of $T^{(L)}_k$. Respectively, every item  $(x^{(L-1)}_{i,v_j}=T^{(L-1)}_{v_j}(\mathbf{x}_i^{(1)}))$ in $R^{(L)}_k(\mathbf{x}_i^{(1)})$ can be further reduced to rule mapping $R^{(L-1)}_{v_j}$ from $(L-2)th$ layer to $(L-1)th$ layer. By replacing every $x^{(L-1)}_{i,v_j}=T^{(L-1)}_{v_j}(\mathbf{x}_i^{(1)})$ to the new mapping, $R^{(L)}_k(\mathbf{x}_i^{(1)})$ can be represented by the mapping from $(L-2)th$ layer to $Lth$ layer. Repeating this process recursively, we can finally get mapping rules from the input space to the $Lth$ output. Specifically, We name this recursive procedure as backward rule induction. Details for that algorithm shown in \textbf{Algorithm \ref{alg:bri}}. 

\begin{algorithm}[htb]
  \caption{Backward Rule Induction}
  \label{alg:bri}
  \begin{algorithmic}[1]
    \REQUIRE
     Input sample $\mathbf{x}^{(1)}_i$; MLD structure $\mathbf{T}$; output derived by \textbf{Algorithm \ref{alg:fdg}} for $\mathbf{x}^{(1)}_i$: $\{T^{(l)}_k(\mathbf{x}^{(1)}_i)$ | $\forall1\le k\le m^{(l)},2\le l\le L\}$; layer index $l$; node index $k$; 
    \ENSURE
      rule mapping $R^{(l)}_k(\mathbf{x}^{(1)}_i)$ from input space to the decision of $kth$ node in $lth$ layer
     
    \STATE {Initialize $R^{(l)}_k(\mathbf{x})=\Phi$ }

      \STATE{Apply \textbf{Algorithm \ref{alg:singlerule}}, input  $(T^{(l-1)}_1(\mathbf{x}^{(1)}_i),...,T^{(l-1)}_{m^{(l-1)}}(\mathbf{x}^{(1)}_i))$ and $T^{(l)}_k$, get $R^{(l)}_k(T^{(l-1)}(\mathbf{x}^{(1)}_i))$ }, assign it to $R^{(l)}_k(\mathbf{x}^{(1)}_i)$
        \FORALL{ rule item $r=(v,c_v,sign)$ in $R^{(l)}_k(\mathbf{x}^{(1)}_i)$}
            
        \STATE{
                Apply \textbf{Algorithm \ref{alg:bri}}, Input $\mathbf{x}^{(1)}_i,\mathbf{T}, \{T^{(l)}_k(\mathbf{x}^{(1)}_i)\}, l-1, v$, get $R^{(l-1)}_v(\mathbf{x}^{(1)}_i)$
            }
            
           \STATE{$R^{(l)}_k(\mathbf{x}^{(1)}_i) = R^{(l)}_k(\mathbf{x}^{(1)}_i)\cup R^{(l-1)}_v(\mathbf{x}^{(1)}_i)$} 
            
            \STATE{Delete $r$ from $R^{(l)}_k(\mathbf{x}^{(1)}_i)$} 
            
        \ENDFOR

  \end{algorithmic}
\end{algorithm}

\begin{figure}[h]
  \centering
  \includegraphics[width=\columnwidth]{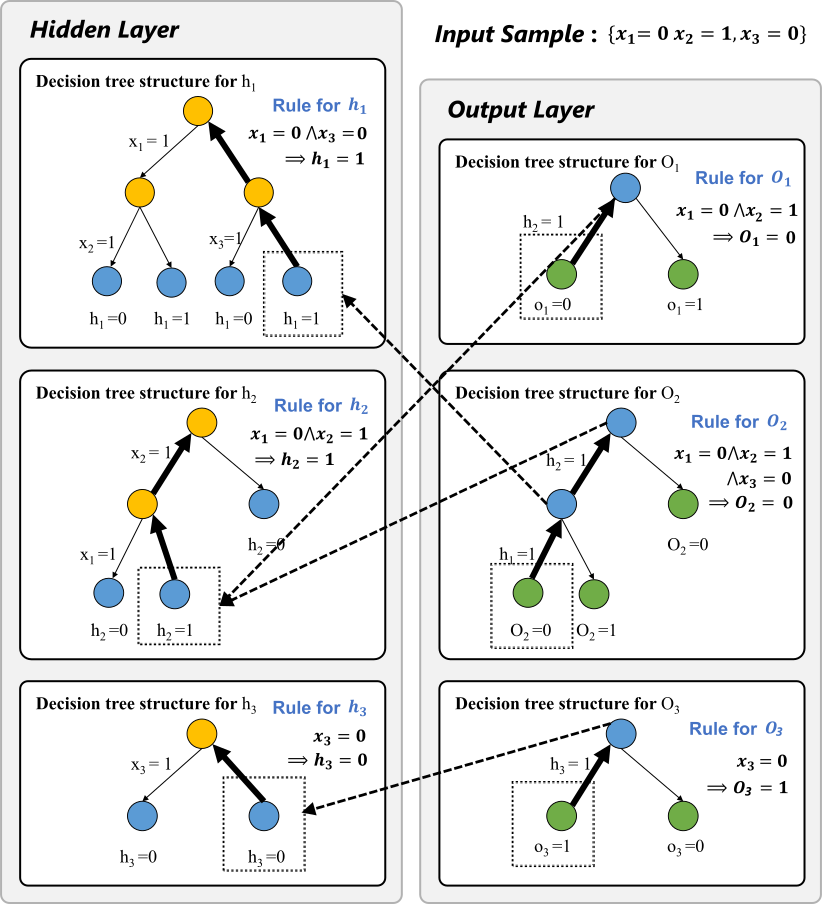}
  \caption{Demonstration for the Backward Rule Induction Process (Target MLP is the structure shown in Figure \ref{fig:discretize})}
\end{figure}

\subsection{Importance Measuring}

In the previous parts, we design the forward rule generation and backward rule induction methods which can be used to derive the local explanation for a given sample. Besides we also want to get some intuitive understanding for the whole model. Importance measuring is one of the common approaches to deliver a global explanation. Here we introduce two importance measure methods based on MLD.

\textbf{Frequency-based method}. On account of the hierarchical structure of MLD, it's hard to measure the feature importance by calculating its information gain as in the ordinary decision tree. But thinking about the construction of decision tree, usually the more sample predictions are given by referring the value of a given feature, the more important this feature is. Inspired by this intuitive idea, an empirical frequency-based method is proposed, in which the importance values are given by calculating how many times a certain feature are used in the reduced decision rule in the sample set: 

\begin{equation}
    VIM^{(Freq)}_v = \sum_{i=1}^n I(v~used~in~Rule(x_i))
\end{equation}

Then we can normalize the importance values:

\begin{equation}
VIM_v^{(Freq)} = \frac{VIM_v^{Freq}}{\sum_{i=1}^d VIM_i^{(Freq)}}
\end{equation}

\textbf{OOB-based method}. Out-of bag method is inspired by the feature measuring in the random forest. The idea of which is first select part of out-of-bag data, then randomly adjust one input feature and observe the impact on the model accuracy. The calculation method is as follow:
\begin{enumerate}
    \item Select a set of out-of-bag data $OOB$, according to the forward decision generation algorithm generate the sample labels based on MLD. Then calculate the errors from the neural network label $errOOB$;
    \item For a given feature $v$, randomly change the feature value of $v$ in $OOB$, calculate the errors between MLD prediction and the neural network labels, denote is as $errOOB_v$;
    \item Calculate the $errOOB_v-errOOB$ as the importance of feature $v$, and normalize it:
    
    \begin{equation}
        VIM^{(OOB)}_v = errOOB_v-errOOB
    \end{equation}
    
    \begin{equation}
    VIM_v^{(OOB)} = \frac{VIM_v^{OOB}}{\sum_{i=1}^d VIM_i^{(OOB)}}
    \end{equation}
    
\end{enumerate}

\section{Experiments}

In this section, we empirically evaluate the effectiveness of the MLD framework on the MNIST and NFPC (National Free Pre-Pregnancy
Check-ups) dataset. After training a multi-layer neural network for each task, we first assess the model predictivity and fidelity of the MLD compared to other pedagogical rule-based methods: decision trees learned using CART, C4.5 and Trepan\cite{craven1996extracting}. Higher predictivity score ensures the explainable model performance more closely matches to the original task, while higher fidelity score indicates higher similarity with the neural network prediction. Second, we provide local and global explanations to give evidence to the interpretability of the MLD framework.  

\subsection{Experimental Setting}

\subsubsection{Dataset}

Our system is evaluated mainly on MNIST and NFPC dataset.  

\begin{itemize}
    \item \textbf{Mnist}: Mnist dataset is a widely used handwritten dataset in the field of machine learning. This dataset contains 60,000 training images and 10,000 testing images. Each of the black and white image is fitted into a 28*28 pixel bounding box.
    \item \textbf{NFPC}: National Free Pre-Pregnancy Check-up(NFPC) is a population-based health survey dataset. This survey mainly focuses on the reproductive-aged couples who wish to conceive, and was conducted across 31 provinces in China from Jan 1, 2014 to Dec 31, 2015. The original data contains detailed personal characteristics of both spouses, which are mainly divided into the following categories: biological indicators (e.g. blood pressure and sugar); personal characteristics of husband and wife (e.g., occupation, education and region); disease characteristics (e.g., Genetic and chronic disease history); personal habits characteristics (e.g., dietary habit and psychology condition). The labels of NFPC data include the indicator of if doctors predict there is a risk of pregnancy and the true fertility outcome.
    For confidentiality reasons, we only use the cleaned and feature-selected data in Yunnan Province to conduct the experiment. The cleaned data contains 106 one-hot features and labels of predicted risk and true outcome.

\end{itemize}

\subsubsection{Baseline}
In terms of predictivity and fidelity, We compare our proposed model with the following pedagogical rule-based interpretability methods:

\begin{itemize}

\item \textbf{C4.5\cite{quinlan2014c4}}: C4.5 is a widely used decision tree construction algorithm, which is an extension of earlier ID3\cite{quinlan1986induction} algorithm. Different from the entropy used in ID3, the splitting criterion of C4.5 is the normalized information gain.

\item \textbf{Cart\cite{breiman2017classification}}: Classification And Regression Tree (CART) can be used in both classification and regression tasks, CART uses Gini index as the splitting criterion.

\item \textbf{Trepan\cite{craven1996extracting}}: Trepan is an inductive method to extract concept description from trained neural network. In the trepan method, a decision tree is learned with queries, while these queries are usually answered by a structure named oracle. The oracle models the data instance and refer the network output to provide answer. 
\end{itemize}

\subsubsection{Network Training}

On MNIST and NFPC dataset, we build two hidden layer MLPs for the target tasks. Mostly 30 nodes in the first hidden layer and 10 nodes in the second layer, while the number of the input nodes and output nodes varies according to the task. On CIFAR-10 dataset, we build a three hidden layer MLP, 128 nodes for the first and second hidden layer and 50 nodes for the third layer. We choose zero mean truncated normal random number to initialize the network parameter and cross entropy loss function as the target loss function. In the training process, we apply adam optimizer and drop out to update the parameters. 

\subsubsection{MLD Construction}

When apply proposed method to extract the MLD structure, we control the size and the depth of the fitted decision trees in order to avoid over-fitting and distill the most representative features as well. Mostly we limit the maximum height to 20 and maximum size to 100 for the fitted decision trees. After fitting process, we apply pruning method to limit the height and size. Specifically, for the multi-class task, we construct two types of decision trees for the output layer. First we train p binary classification decision trees to produce interpretation for each class separately (p is the number of classes), Second we train a single p classification decision tree to measure the performance. 

\begin{table*}
  \caption{Predictivity and Fidelity Measure}
  \label{tab:compare}
  \begin{tabular}{cccccccccc}
    \toprule
    \multicolumn{2}{c}{}& \multicolumn{4}{c}{Predictivity Measure} &\multicolumn{4}{c}{Fidelity Measure} \\
    \cmidrule(lr){3-6}\cmidrule(lr){7-10}
   Dataset & Method & Acc(Train)&	F1(Train)&	Acc(Test)&	F1(Test) & Acc(Train)&	F1(Train)&	Acc(Test)&	F1(Test) \\
    \midrule
   Discretized MNIST & Target MLP & 0.9622 & 0.9684 & 0.9652 & 0.9545 &-&-&-&- \\
   &\textbf{MLD}& \textbf{0.9025} & \textbf{0.8943} & \textbf{0.8976} & \textbf{0.9034} & \textbf{0.8936} & \textbf{0.9034} & \textbf{0.8950} &\textbf{0.8859}\\
   &CART&0.8332& 0.8345 & 0.8381 & 0.8471 & 0.8589 & 0.8364 & 0.8459& 0.8427 \\
   &C4.5&0.8329&0.8359&0.8432&0.8324&0.8531&0.8642&0.8434&0.8531\\ 
   &Trepan&0.8675&0.8794&0.8689&0.8845&0.8788&0.8710&0.8835&0.8753\\
   \midrule
   NFPC(Predicted Risk) & Target MLP &0.9945&0.9873&0.9939&0.9926&-&-&-&-\\
   &\textbf{MLD}&\textbf{0.9854}&0.9768&0.9820&\textbf{0.9875}&\textbf{0.9945}&\textbf{0.9920}&\textbf{0.9898}&\textbf{0.9854}\\
   &CART&0.9754&0.9660&0.9659&0.9735&0.9749&0.9750&0.9850&0.9829\\
   &C4.5&0.9740&0.9758&0.9670&0.9720&0.9830&0.9734&0.9819&0.9783\\
   &Trepan&0.9845&\textbf{0.9853}&\textbf{0.9950}&0.9898&0.9929&0.9830&0.9914&0.9840\\
   \midrule
   NFPC(True Outcome) & Target MLP &0.9854&0.9834&0.9935&0.9845&-&-&-&-\\
   &\textbf{MLD}&0.9678&0.9723&\textbf{0.9846}&\textbf{0.9743}&\textbf{0.9874}&0.9734&0.9774&0.9758\\
   &CART&0.9618&\textbf{0.9732}&0.9691&0.9745&0.9812&\textbf{0.9829}&0.9749&\textbf{0.9764}\\
   &C4.5&0.9628&0.9723&0.9634&0.9712&0.9718&0.9814&0.9712&0.9714\\
   &Trepan&\textbf{0.9745}&0.9634&0.9745&0.9684&0.9734&0.9812&\textbf{0.9842}&0.9674\\
   \midrule
   CIFAR10 & Target MLP & 0.4820&0.4832&0.4880&0.5023&-&-&-&-\\
   &\textbf{MLD}&\textbf{0.3125}&\textbf{0.3251}&\textbf{0.3278}&\textbf{0.3095}&\textbf{0.3192}&\textbf{0.3150}&\textbf{0.3144}&\textbf{0.3125}\\
   &CART&0.2765&0.2845&0.2940&0.2554&0.2689&0.2834&0.2525&0.2678\\
   &C4.5&0.2637&0.2832&0.2734&0.2934&0.2714&0.2697&0.2813&0.2664\\
   &Trepan&0.2923&0.3023&0.2912&0.3038&0.2914&0.2943&0.2834&0.3060\\
  \bottomrule
\end{tabular}
\end{table*}

\subsection{Predictivity and Fidelity Measure}

 One of the approaches to assess the goodness of our proposed framework is by measuring the predictivity and fidelity. Where the predictivity score shows how good the extracted model fits the   real data labels, while the fidelity score indicates how good the extracted model fits the labels predicted by the trained neural network model. 
 More precisely, apply the forward rule generation algorithm, we can reproduce data labels  based on the MLD. Then we measure the predictivity and fidelity in terms of four scores: accuray and F1 score in train and test dataset. Higher predictivity scores guarantee the extracted model matches the origin data well, but achieving high fidelity scores may seems more important, since it ensures the extracted model actually obtains the insights into the neural network model.
 As shown in \textbf{Table \ref{tab:compare}}, we compare our proposed MLD framework with several baselines over several datasets. Most of the results show our model overperforms over other baseline methods.

\begin{figure}[t]
  \centering
  \subfigure[MNIST Sample Case For Digit 8]{
    \includegraphics[width=\columnwidth]{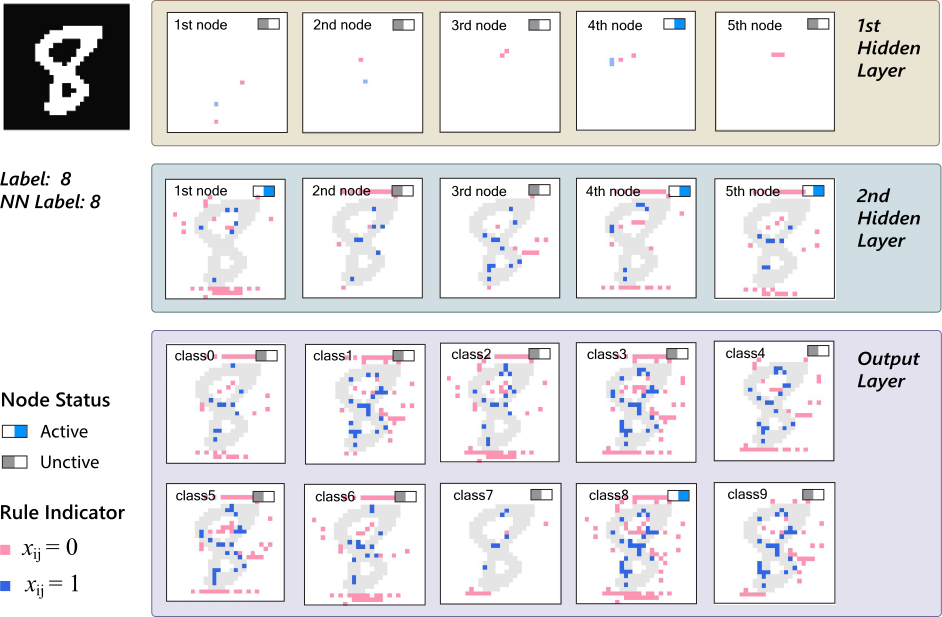}
  }
  \subfigure[MNIST Sample Case For Digit 4]{
    \includegraphics[width=\columnwidth]{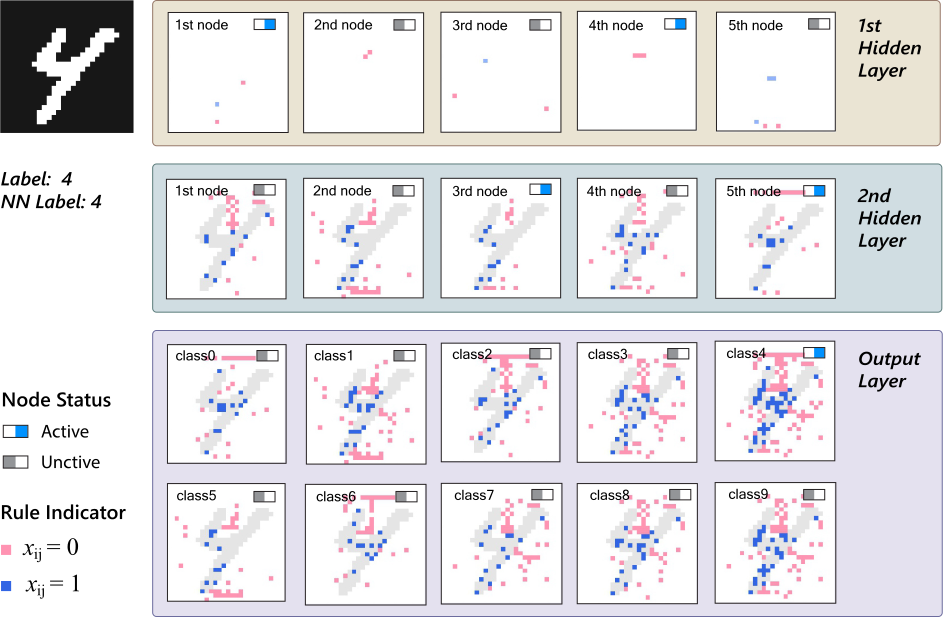}
  }
  \subfigure[MNIST Sample Case For Digit 2 (misclassification)]{
  \includegraphics[width=\columnwidth]{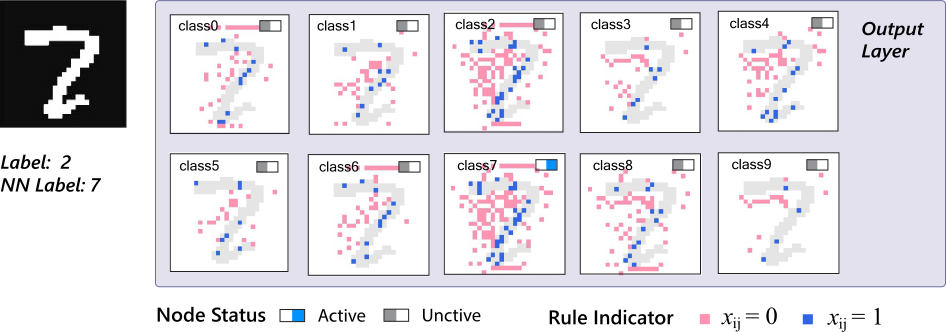}
  }
  \caption{Local Explanation Cases for MNIST}
  \label{fig:cases}
\end{figure}

\subsection{Interpretability for MNIST Dataset}

In terms of interpretability, our goal for the MNIST task is extracting visual explanations of the model decision in sample and class scales. Note that, in order to generate explicit visual explanation, we first descretize all the input value of MNIST dataset into 0 and 1. 

\subsubsection{Local Explanation Cases}
In the MLD framework, we can apply backward rule induction method to extract decision rules of the neural network regarding to a given sample. In the example for digit 8 given in \textbf{Figure \ref{fig:cases}}, the real label and neural network decision are the same. In the result of MLD, it is also classified correctly as 8 and the other incorrect classifications are rejected as well. Observing this visual example, it's easy to tell that the derived rules become more and more complicate as the layer increases. This result comes from the nature of the MLD structure: except for the input layer, sample rules for nodes in each layer are the combination of the rules derived in the previous layer. Since the MLD is directly generated based on the neural network, observing how the MLD works may help us better understand the underlying working principle of neural networks.
 
Observing the rules extracted in the output layer for digit 8 in \textbf{Figure \ref{fig:cases}}: the number of rules to classify it to digit 8 correctly is the most, the numbers of the rules which refuse to classify it as digit 3, 5, 9 are relatively larger compare to other digits. Intuitively, shape of number 8 is indeed closer to 3, 5, 9 than other numbers, therefore more evidence is needed to distinguish it from 3, 5, 9. From the example of digit 4 sample given in \textbf{Figure \ref{fig:cases}}, we can also discover the similar pattern: shape of number 4 is much closer to number 9, thus the number of rules required to correctly identify it as 4 is the most, while the number of rules refuse to recognize it as 9 is the second largest.

The local explanation given by the MLD framework may also give us some intuition about why the neural network sometimes doesn't provide the right decision. In the case for digit 2 shown in \textbf{Figure \ref{fig:cases}}, the neural network wrongly classified it as digit 7. Through the visual explanation, we can roughly tell the reason for such misclassification: the neural network may not take the small tail of the digit 2 into account in the classification process.

\begin{figure}[t]
  \centering
  \subfigure[Positive Rule Importance Distribution for Each Classification]{
    \includegraphics[width=\columnwidth]{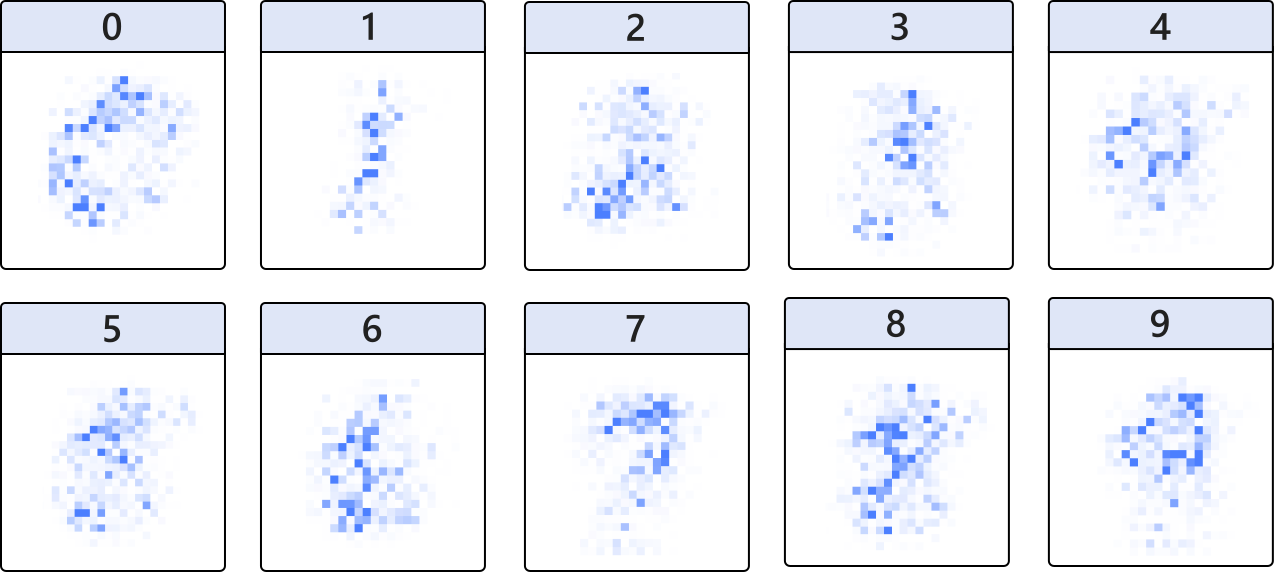}
  }
  \subfigure[Negative Rule Importance Distribution for Each Classification]{
    \includegraphics[width=\columnwidth]{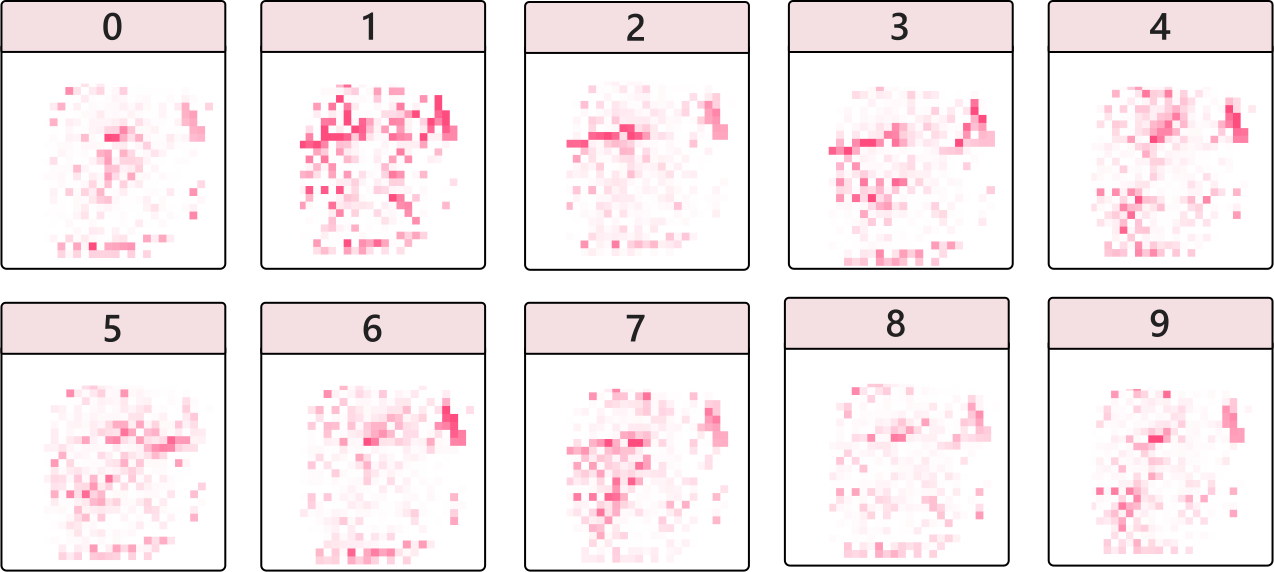}
  }
  \caption{Rule Importance Distribution for Each Classification on MNIST}
  \label{fig:mnistclass}
\end{figure}

\subsubsection{Global Explanation}

In terms of global explanation, apply the frequency-based method proposed in the MLD framework, we can obtain the importance score for each rule related to the input feature. Specifically, in the discretized MNIST scenario, for each of the ten classifications, rules correspond with each input pixel like $\{x^{(1)}_{1} = 1 \},\{x^{(1)}_{3} = 0 \} $ can be assigned to a score according to how many times a certain rule is used to generate sample decisions. The visualized results of the global explanation are shown in \textbf{Figure \ref{fig:mnistclass}}, in which the importance scores for positive and negative rules are mapped in the 28*28 pixels. From this distribution we can roughly tell the outline and periphery of the ten numbers. This intuitive result also suggests that the global explanation given by our MLD framework is comprehensive and reasonable.

\subsection{Interpretability for NFPC Data}

Our goal for the interpretability in NFPC dataset is mainly focus on providing global explanation and assessing the explanation as well. More specifically, first we train MLP for doctor predicted risk and the true outcome task. Then we construct MLD and apply the proposed importance measure method to extract top-10 important features for the two tasks separately. Furthermore for comparison, we train random forest model for the two tasks and get another two sets of top-10 importance measure. Finally we compare the importance measures given by MLD and random forest by fitting the target task again and performing human experiment.

\subsubsection{Compare the Explanations By Regression}

In order to measure how representative the extracted features are and conduct comparison, We use the top-10 features extracted from the MLD and random forest in both of the tasks to fit logistic regression models separately, then we compare the AUC score. The result is shown in \textbf{Table \ref{tab:logitregression}}. From the result we can see though the scores of MLD are slightly lower than random forest, there's no significant difference between the two model. That means the top-10 key features selected by MLD and random forest have almost the same explanatory power.

\begin{table}
  \caption{Logistic Regression Fit Results Using Top-10 Important Features}
  \label{tab:logitregression}
  \begin{tabular}{ccc}
    \toprule
    Task & AUC(MLD) & AUC(RF) \\
    \midrule
    Predicted Risk&0.7245&0.7329\\
    \midrule
    True Outcome&0.6628&0.6684\\
    \bottomrule
  \end{tabular}
\end{table}

\subsubsection{Compare the Explanations By Human Experiment}

One of the most important aspects for the interpretability is providing the explanations which can be understood by human beings, thus performing human experiment is necessary for assessing the interpretability methods. In the NFPC case, we conduct a user study to compare the key feature sets derived from MLD and random forest model. We interviewed 40 
obstetricians through the internet, each participant was given one set of top-10 important features for predicited risk task and one set of top-10 important features for true outcome task randomly (randomization here means each set can either from the result given by MLD or random forest, but make sure each of the set is assigned for 20 obstetricians). Then they were asked to score each feature from 0 to 3 according to how much they think this feature has an impact on fertility outcomes. 
The reference scoring criteria is as follows: 0-no effect, 1-little effect, 2-moderate effect, 3-significant effect. After that, we sum up the scores for each set given by each obstetrician and then perform paired t-test to compare the difference. Related Results are given in \textbf{Table \ref{tab:ttest}}.

\begin{table}
  \caption{ User Study Paired One-Sided t-Test Result}
  \label{tab:ttest}
  \begin{tabular}{ccccc}
    \toprule
    Task & $\mu_{score}$(MLD) & $\mu_{score}$(RF) &  t & p (>t)\\
    \midrule
    Predicted Risk& 20.6 & 20.8 & 1.2666 & 0.22\\
    \midrule
    True Outcome& 10.5 & 8.6 & 24.06& 0\\
  \bottomrule
\end{tabular}
\end{table}

If we choose the 95\% confidence interval, the p value for predicted risk task is 0.22, for true outcome is 0, indicates regarding to the total set score, for the predicted risk task there's no significant difference between the key feature sets extracted by MLD and random forest. While for the true outcome task, the MLD scores are significant higher than random forest scores.

\section{Conclusion}

In this paper, we propose a multi-level decision (MLD) framework to generate explanation for the multi-layer neural network model. At first, a multi-level decision structure is built to reconstruct the original neural network. Based on the MLD structure, forward decision generation algorithm and backward rule induction algorithm are applied to derive the sample decision and sample rule mapping, frequency-based and OOB-based method are applied to measure the feature importance. Furthermore, functional-grounded and human-grounded experiments are also carried out to evaluate the interpretability.

\bibliographystyle{ACM-Reference-Format}
\bibliography{ref}

\end{document}